\documentclass[letterpaper]{article} 
\usepackage[]{aaai23}  
\usepackage{times}  
\usepackage{helvet}  
\usepackage{courier}  
\usepackage[hyphens]{url}  
\usepackage{graphicx} 
\usepackage{natbib}  
\usepackage{caption} 
\usepackage{algorithm}
\usepackage{algorithmic}

\usepackage{newfloat}
\usepackage{listings}
\DeclareCaptionStyle{ruled}{labelfont=normalfont,labelsep=colon,strut=off} 
\floatstyle{ruled}
\newfloat{listing}{tb}{lst}{}
\floatname{listing}{Listing}
 \nocopyright 

\title{LUCID: Exposing Algorithmic Bias through Inverse Design}

\author {
    Carmen Mazijn,\footnote{Data Lab, VUB -- carmen.mazijn@vub.be}
    Carina Prunkl,\footnote{Institute for Ethics in AI, Oxford University\\ carina.prunkl@philosophy.ox.ac.uk}
    Andres Algaba, \footnote{Data Lab, VUB -- andres.algaba@vub.be}
    Jan Danckaert, \footnote{APHY, VUB -- jan.danckaert@vub.be}
    Vincent Ginis \footnote{Data Lab, VUB; SEAS, Harvard -- vincent.ginis@vub.be}
}

\begin{document}

\maketitle

\begin{abstract}
AI systems can create, propagate, support, and automate bias in decision-making processes. To mitigate biased decisions, we both need to understand the origin of the bias and define what it means for an algorithm to make fair decisions. Most group fairness notions assess a model's equality of outcome by computing statistical metrics on the outputs. We argue that these output metrics encounter intrinsic obstacles and present a complementary approach that aligns with the increasing focus on equality of treatment. By Locating Unfairness through Canonical Inverse Design (LUCID), we generate a canonical set that shows the desired inputs for a model given a preferred output. The canonical set reveals the model's internal logic and exposes potential unethical biases by repeatedly interrogating the decision-making process. We evaluate LUCID on the UCI Adult and COMPAS data sets and find that some biases detected by a canonical set differ from those of output metrics. The results show that by shifting the focus towards equality of treatment and looking into the algorithm's internal workings, the canonical sets are a valuable addition to the toolbox of algorithmic fairness evaluation.
\end{abstract}

\section{Introduction}
\label{sec:intro}
Artificial intelligence (AI) systems are used in decision-making processes throughout all aspects of human life, ranging from detecting child abuse, determining access to education or healthcare, and granting loans~\citep{amrit2017identifying,ledford2019millions,makhlouf2021applicability}. However, it is by now a well-established fact that algorithms can be biased and lead to discrimination against already disadvantaged population groups \citep{barocas2016big,buolamwini2018gender,chouldechova2018frontiers,whittaker2018ai}. The sources of such biases are multiple and include problem specification, historical bias, unrepresentative data, biased measurement methods, or choice of objective function \citep{barocas2019,fazelpour2021algorithmic,lee2021risk,suresh2021framework}.

\begin{figure*}[t!]
\centering
\includegraphics[width=0.90\textwidth]{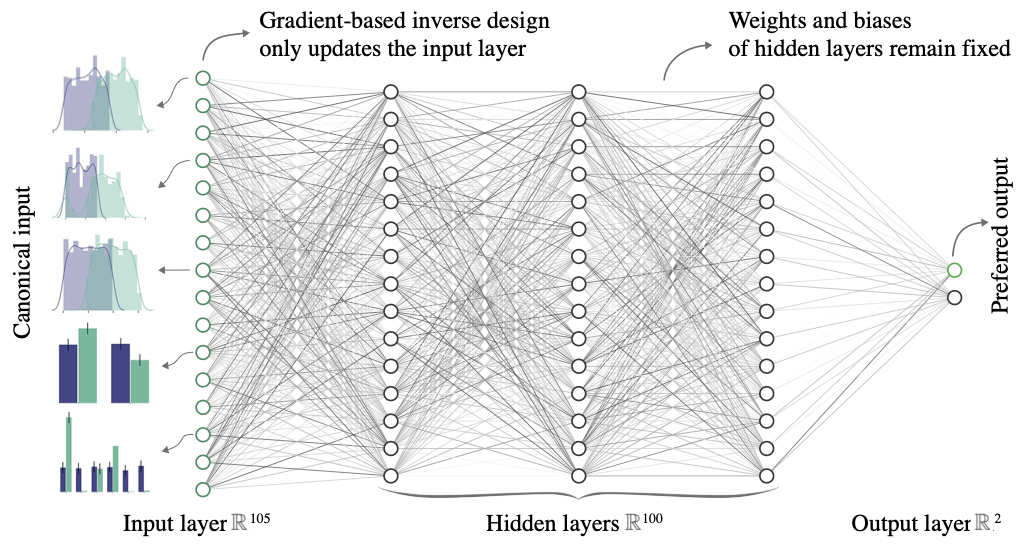}
\caption{An overview of LUCID for a binary classifier trained on the UCI Adult data set. Through gradient-based inverse design on the input layer, we generate a canonical set for a preferred output. The weights and biases of the model remain fixed; the greyscale of each connection encodes the fixed value. The canonical set reveals the model's internal logic and is visualized via the histograms. We locate the unfairness by analyzing whether the distributions of the protected features within the canonical set remain balanced after inverse design.}
\label{fig:inversedesign}
\end{figure*}

Recent efforts to identify algorithmic discrimination often focus on the statistical properties of a model's \textit{output}. The standard approach is to translate philosophical or political notions of group fairness into a statistical metric~\cite{makhlouf2021applicability}. The model's output can then be analysed with respect to the chosen notion of group fairness and the model is judged to be ``fair'' or ``unfair''. There are several widely recognised issues with output-based fairness evaluations of this kind. First, there often is substantial philosophical disagreement as to what ought to be considered a ``fair'' outcome distribution~\cite{binns2018fairness,gallie2019essentially}. The now infamous controversy about the alleged racism of the COMPAS recidivism risk algorithm boiled down to such a disagreement. In this case, the two fairness measures under debate were accuracy equality and equalised odds with respect to race. Second, different notions of group fairness are incompatible with each other, except for highly special cases \cite{kleinberg2016inherent}. Third, the computation of fairness metrics depends on a benchmark data set, which is often also used to evaluate the model on other metrics such as accuracy. The metrics do not assess the model's fairness towards the whole population after deployment~\cite{northcutt2021pervasive}. Fourth, work on group fairness usually relies on the evaluation of a limited number of prescribed protected attributes, running the risk of missing discrimination either against people who are at the intersection of different groups or against groups that do not share a protected characteristic~\citep{binns2020apparent,crenshaw1990mapping,wachter2019right}. Finally, focussing exclusively on output distributions to determine fairness is only one part of the story. In everyday life and when stakes are high, the output of a given decision is not the only thing one typically cares about. We are also interested in how the decision came about, e.g. \textit{why} I wasn't granted the loan I applied for or \textit{why} I didn't receive the job I interviewed for~\cite{AIHLE2019,wachter2017counterfactual}. Understanding the reasoning behind a decision is not just relevant from a moral point of view, it is equally important within a legal context. Disparate treatment and direct discrimination both aim at addressing cases in which similarly situated individuals are not treated alike on grounds of a legally protected characteristic. In these cases, it becomes relevant both \textit{that} such individuals were treated differently and \textit{why} they were treated differently. Output-based fairness evaluations cannot address these issues as they do not take into account the internal logic of the model in question.

We present a method (called LUCID, for Locating Unfairness through Canonical Inverse Design), complementary to output-based fairness evaluations, that takes into account a model's internal logic. In particular, we introduce the notion of a ``canonical set'' that allows us to evaluate the fairness of a model's decision-making processes (see Fig.~\ref{fig:inversedesign}). Through gradient-based inverse design, we generate a canonical input, which can be thought of as the desired input given a preferred output for a trained model. The canonical set then emerges from repeatedly interrogating the model's decision-making process by generating canonical inputs. By revealing information about the model's mechanisms, the canonical set provides insights into how the model reaches certain decisions, e.g., what features play a crucial role in the model's decision-making process. To locate unfairness in the model's logic, we inspect the distribution of a protected demographic feature within the canonical set. This approach aligns with the increasing focus on equality of treatment beyond equality of outcome.

In contrast to traditional fairness evaluations, LUCID does not require a specific fairness metric, a ground truth, or a benchmark data set. We show that LUCID can be applied to any differentiable model. 
As an illustration, we evaluate binary classifiers trained on the UCI Adult~\cite{dua2019adult} and COMPAS~\cite{angwin2016compas} data sets. We find that analyzing the canonical set exposes several unethical biases, which interestingly differ from those found by traditional group fairness metrics. By looking into the algorithm's internal workings, the canonical sets are a valuable addition when evaluating algorithmic fairness.

\section{Background and Related Work}
\label{sec:literature}

While, technically, many countries have anti-discrimination laws in place that are designed to protect people from discriminatory harms~\cite{barocas2017fairness,GDPReu}, there are often considerable practical difficulties associated with the very detection of algorithmic discrimination. The first problem is that those subject to unfair treatment (e.g., the rejected job applicant) often lack the epistemic resources needed to identify instances of discrimination~\cite{milano2021epistemic,selbst2018intuitive}. The second problem is that access to data sets by third parties is often limited due to intellectual property and privacy rights. Finally, the third problem relates to the frequent opaqueness of the algorithmic decision-making process itself. LUCID mainly addresses the second and third problems.

LUCID lies at the intersection of fairness and interpretability in algorithmic decision-making. There is a strong interaction between these fields, and their connections~\cite{meng2022mimic} and trade-offs~\cite{kleinberg2019simple} are part of ongoing research. The most widespread group of methods to gauge fair decision-making translate philosophical or political notions of group fairness into mathematical statements on the model's output~\cite{makhlouf2021applicability}. The number of this kind of fairness metrics has grown over the past years, accounting for at least $19$ definitions~\cite{barocas2019,hardt2016equality,zafar2017fairness}. Furthermore, most prominent open-source fairness toolkits rely on these statistical output metrics~\cite{lee2021landscape}.

Over the past few years, much work has been done on post hoc interpretability methods, especially in the computer vision literature~\cite{das2020opportunities}. The most prominent examples are feature importance estimation methods that help understand which features have a high impact on the output of a model by giving a score to each input. While the feature importance estimation methods differ in various ways, they can be broadly categorized into perturbation- and gradient-based explanations~\cite{agarwal2021unification}. 

SHAP~\cite{lundberg2017unified} is an example of the former as it constructs local explanations of decision-making algorithms by using perturbations of individual samples. However, the resulting explanations are found to be unreliable, especially in the context of fairness~\cite{slack2020fooling,balagopalan2022explain}. Moreover, the standard implementation method does not account for feature dependence whereas many other implementations are computationally very expensive~\cite{gohel2021explainable}. Nevertheless, perturbation-based methods are often used in combination with statistical fairness metrics~\cite{datta2016qii}. Thereby, this interpretable fairness analysis inherits the obstacles from the output metrics, namely philosophical disagreements~\cite{binns2018fairness,gallie2019essentially}, statistical incompatibilities~\cite{kleinberg2016inherent}, the absence of universal ground truth, and the selection of a benchmark data set~\cite{northcutt2021pervasive} and a limited number of prescribed protected attributes~\citep{binns2020apparent}.

The Integrated Gradients~\cite{sundararajan2017attribution} method is an example of gradient-based explanations. The gradients of the outputs of individual samples with respect to their inputs are used to construct local explanations. Importantly, the gradients can also be used to construct global explanations by generating an input with the highest activation and certainty for a specific output starting from random noise~\cite{simonyan2014deep}. One of the first implementations of this method, called DeepDream, was developed at Google~\cite{Deepdream} in $2015$. The resulting input images illustrate which elements are essential to get the preferred output. For example, when generating input images that optimize the output ``dog,'' the results are images with many dog legs, ears, and noses, merged in an unnatural, almost psychedelic way. The resulting images are therefore called ``dreams.''

Since then, the technique to determine hidden parameters of a complex system through inverse design has been used in several other research fields, including physics, computer science, engineering, and biotechnology~\cite{ferruz2022dreaming,forte2022inverse,lenaerts2021artificial,ren2020benchmarking}. We show that LUCID builds upon these methods as the canonical sets are the result of repeatedly applying inverse design to generate canonical inputs, thereby revealing the internal logic of a trained model.

\section{Canonical Inverse Design}
\label{sec:methodology}
Conventional neural networks use gradient descent to improve their workings by taking advantage of their mathematical structure, which can be differentiated straightforwardly~\cite{nielsen2015neural}. All the layers in a model can be optimized through gradient descent, including the input values. Indeed, the input vector can be seen as a special layer of the model. With LUCID, we use this property to create a canonical input for a preferred output. In other words, starting from a random input vector one can construct the ideal input of a trained model through gradient descent on the input layer, keeping the weights and biases fixed.

This gradient-based inverse design has been extensively used in the computer vision literature~\cite{Deepdream,simonyan2014deep,sundararajan2017attribution}, but there is a key difference in our application to tabular data. For images, canonical inputs are interpretable individually and difficult to aggregate, whereas for tabular data we have the opposite scenario. In addition, due to the stochastic nature of randomly generated vectors, there is little information in the canonical version of a single input vector. Therefore, we generate a canonical set which results from repeatedly interrogating the model's decision-making process by generating canonical inputs, revealing its internal workings. To locate unfairness in the model's logic, we inspect the distribution of a protected demographic feature within the canonical set, and compare it to the initial random distribution. This approach aligns with the increasing focus on equality of treatment beyond equality of outcome, as this requires interpretability, which builds and supports trust, and contributes to procedural fairness.

In Algorithm~\ref{alg:algo}, we show an implementation of LUCID and detail how a canonical set can be generated for a differentiable model by updating a random input vector via gradient descent on the input layer. First, we generate an extensive set of randomly initialized input vectors where the features are sampled from a uniform distribution. Then, these randomly initialized input vectors are transformed into canonical inputs through inverse design. The transformations are the result of minimizing the loss between the model predictions and the preferred output (e.g., a loan is granted). Afterwards, the canonical set is analyzed to learn about the internal workings of the model and evaluated if the model is insensitive to protected attributes.

\begin{algorithm}[t]
\caption{LUCID, our proposed algorithm. The default setup in our paper: number of canonical inputs $N = 1000$, number of epochs $E = 200$, learning rate $\alpha = 0.1$, a binary classifier $s$, and a cross-entropy loss function $l$.}
\label{alg:algo}
\textbf{Inputs}: A differentiable model ($s(X)$), and objective function ($l(\hat{y}|y)$) with prediction $\hat{y}$ and preferred output $y$.\\
\textbf{Parameters}: Number of canonical inputs ($N$), epochs ($E$), and learning rate ($\alpha$).\\
\textbf{Output}: A canonical set ($X$).
\begin{algorithmic}[1]
\STATE $M \gets length(X)$
\FOR{$i = 0, ..., N$}
\STATE $\{X_i^{(m)}\}_{m=1}^M \sim \mathcal{U}(0,1)$
\FOR{$j = 0, ..., E$}
\STATE $\hat{y}_j \gets s(X_i)$
\STATE $X_i \gets X_i - \alpha \nabla_{X_i} l(\hat{y}_j|y)$ 
\ENDFOR
\ENDFOR
\RETURN $X$
\end{algorithmic}
\end{algorithm}

\subsection{Design Considerations}
We discuss the details of how to implement LUCID to construct the canonical set, including the encoding of features into vectors, the initialization of the vectors, the relation between the learning rate and the number of epochs, and the specification of the preferred output. While these choices have an influence on the resulting canonical set, the method itself is agnostic to these choices.

\subsubsection*{Numerical vs. Categorical Features}
Data referring to humans usually contains categorical features such as gender, occupation, and nationality, and numerical features such as age, weight, and income. However, models only process numerical inputs. To feed data to a neural network, the categorical features are encoded. These techniques include ``one-hot encoding'' if the number of categories is known when designing the model, ``hash encoding'' if the number of categories is not known upfront, ``label encoding'' to transform a categorical feature into numerical values~\cite{encoding21}.

\subsubsection*{Initialization of the Input Vectors}
The input layer needs to be initialized with a set of randomly generated vectors. These vectors can be created in multiple ways. Indeed, the features in the training data each satisfy a particular distribution. These distributions might be the result of defective data collection practices, might not represent the distributions of the populations, or reveal the impact of discriminatory practices. To create the initial input vectors, the values of the numerical and categorical features could follow these distributions or they could be drawn from a random distribution. The latter option ensures an entirely random initialization and also works when the training data set is unknown. 

\begin{figure}[t]
    \centering
    \includegraphics[width=0.90\columnwidth]{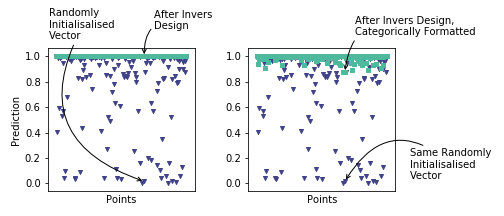}
    \caption{The left panel shows the model's predictions of randomly generated vectors before (dark blue) and after (light green) being updated through gradient-based inverse design with a learning rate of $0.5$. In the right panel, the canonical inputs are categorically formatted and again predicted by the model. The prediction is expressed as a value between $0$ and $1$ with a threshold of $0.5$. There is a small loss of information due to the categorical formatting, but all vectors are still positively predicted. The results in these plots were generated by LUCID for a binary classifier trained on the UCI Adult data set.}
    \label{fig:preds}
\end{figure}

The choice of the random distribution and its parameters depend on the range of realistic values for the inputs. For example, a uniform distribution between zero and one is a valid choice if the features are all between zero and one after pre-processing with min-max scaling. Finally, before starting the inverse design process, the categorical features of each input vector can be formatted to correspond to a ``real'' sample. For example, the categorical features can be one-hot encoded by assigning the category with the largest value a numerical value of one, and setting all the other categories to zero. See Fig.~\ref{fig:preds} for the distribution of the initialized vectors (in dark blue) when the numerical and categorical features are generated according to a uniform distribution for the binary classifier trained on the UCI Adult data set.

\subsubsection*{Evolution of Numerical Features}
In practice, numerical characteristics typically have lower and upper limits. The inverse design process is agnostic to the meaning of these values and might update them outside of the real-world range. It is possible to enforce chosen boundaries in each epoch, each couple of epochs, or simply at the end. Enforcing boundaries shift the focus to update other features. This also means that certain information is lost. Therefore, we do not limit the numerical features as the shift of these values contains information.

\begin{figure*}[t]
    \centering
    \includegraphics[width=0.90\textwidth]{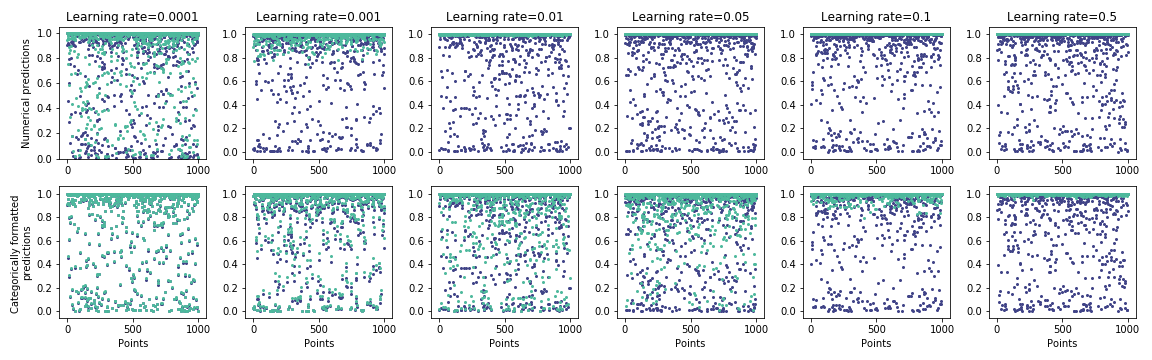}
    \caption{The learning rate parameter of inverse design influences the generation of the canonical inputs. For six different learning rates between $10^{-4}$ and $0.5$, a thousand vectors are randomly initialized following a uniform distribution. On the first row, the predictions are plotted before (dark blue) and after (light green) inverse design. On the second row, the predictions of the same initial thousand vectors and the canonical inputs after categorical formatting are plotted. The canonical inputs receive a higher prediction when the learning rate is higher. After categorical formatting, the predictions are lower due to a loss of information. Each inverse design has the same amount of epochs, here $200$. The results in these plots were generated by LUCID for the UCI Adult classifier.}
    \label{fig:lrs}
\end{figure*}

\begin{figure*}[t]
\centering
\includegraphics[width=0.90\textwidth]{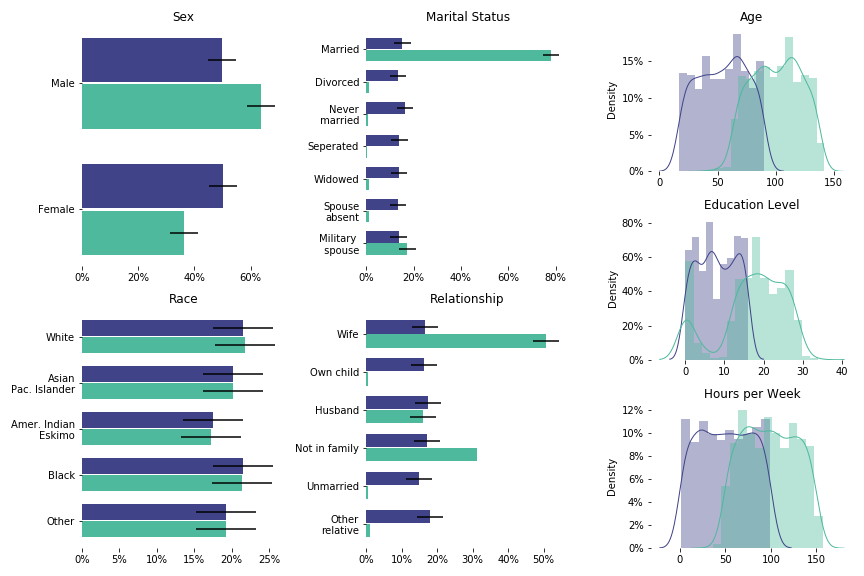}
\caption{To locate unfairness in the binary classifier trained on the UCI Adult data set, we apply LUCID to assess if the protected features have a uniform distribution in the canonical set. This canonical set was created with a learning rate of $0.1$ looping over $200$ epochs. To assess if it is balanced w.r.t. the four protected features ``sex,'' ``race,'' ``marital status,'' ``relationship,'' their distribution before (dark blue) and after (light green) inverse design is analyzed. The error bars indicate the variance of the distribution with the respective number of categories. We learn that the ``race'' feature keeps its uniform distribution. ``Sex,'' ``marital status,'' and ``relationship'' do not keep their uniform distribution after inverse design. This indicates a preference of the model for certain values of these features. Additionally, three numerical features are analyzed: ``age,'' ``education level,'' and ``hours per week.'' All three distributions shift to higher values to achieve a positive output.}
\label{fig:interpret1}
\end{figure*}

\subsubsection*{Categorical Formatting}
When a categorical feature is one-hot-encoded, the vector includes zeros for the number of possible values. One of these entries then receives the value one, signaling the category of the feature. However, during the inverse design process, all values in the input vector are updated. This process, therefore, can also update all the values which were initially zero. However, a vector with real values on all positions does not correspond to an actual sample, with only one specific value for each feature. Therefore we need to format those vectors to correspond to ``real'' samples. For each categorical feature, the highest value related to said feature indicates the value of the category and is indicated as one. All the other positions reset to zero. Note that a part of the information in the vector is now lost. The impact on the predictions of the numerical vectors and the formatted vectors is shown between the first and second row in Fig.~\ref{fig:lrs} for the neural network trained on the UCI Adult data set. It is possible to format the input vector during each couple of epoch(s), or only at the end during the inverse design process. We have chosen the latter for this paper.

\subsubsection*{Learning Rate and Number of Epochs}
The traditional gradient descent method requires a ``learning rate'' which determines how much the weights and biases are adjusted in each epoch. During the learning phase, this parameter varies typically between $10^{-4}$ and $10^{-1}$~\cite{nielsen2015neural}. In LUCID, the learning rate indicates how fast the randomly generated input vectors are adjusted. It can typically take larger values as only the input layer is being updated, while during training all weights and biases are updated.

The aim of generating the canonical set is that all inputs strongly activate the preferred output. The learning rate and the number of epochs both have an influence on how fast this is achieved. See Fig.~\ref{fig:lrs} for the evolution of the predictions when the learning rate increases for the binary classifier trained on the UCI Adult data set. A sufficient number of epochs is needed to achieve adequate vectors in the canonical set. When the learning rate is lower, the number of epochs needs to be higher to achieve a canonical set with a similar mean loss. However, the categorical features may not often change when updating the vector for very small learning rates.

\subsubsection*{The Preferred Output}
A canonical set can be generated for each possible output. The two most common tasks are classification and regression. In the former, the output is often a one-hot encoded vector with one (or multiple) input(s) representing the relevant category (or categories in the case of a multiclass or -label task). For regression, the output is often a scalar or vector with real values. Each of these task has a corresponding loss function that can be minimized, the most popular being cross-entropy for classification and mean squared errors for regression. Note that as long as the models are differentiable LUCID can be applied. The preferred output  depends on the model’s task and the evaluation of the fairness question. The positive (negative) output results in a (dis)advantage for the individual. The canonical set of this output tells us which features positively (negatively) impact the decision-making process.In this paper, we focus on the positive outputs and the classification tasks which allows us to better compare the canonical sets with the output-based fairness metrics.

\section{Locating Unfairness}
\label{sec:results}
We evaluate LUCID on the UCI Adult~\cite{dua2019adult} and COMPAS~\cite{angwin2016compas} data sets. The task of UCI Adult is to predict if a person earns more or less than $\$50,000$ per year. The preferred outcome is that a person earns more. We focus on the legally protected featured encoded in the UCI Adult data set as ``sex,'' ``race,'' ``native country,'' ``marital status,'' and ``relationship.'' For COMPAS, the task is to predict if a person will commit recidivism in the next two years. The preferred outcome is that a person is predicted to not commit recidivism and can thus be released on bail. The legally protected features in the COMPAS data set are ``race,'' and ``sex.'' The models for both tasks are binary fully-connected neural network classifiers. The classifiers consist of hidden layers with ReLU activation functions and a softmax output layer. The number of nodes and layers is decided by the accuracy on the test set. We choose for this rather simple and small architecture as the point of this experiment is not to achieve state--of--the--art performance, but to demonstrate the capabilities of LUCID and compare it with output--based fairness metrics. Note that the computational complexity of LUCID depends on the model, and the number of epochs and inputs.

\begin{figure}[t]
    \centering
    \includegraphics[width=0.90\columnwidth]{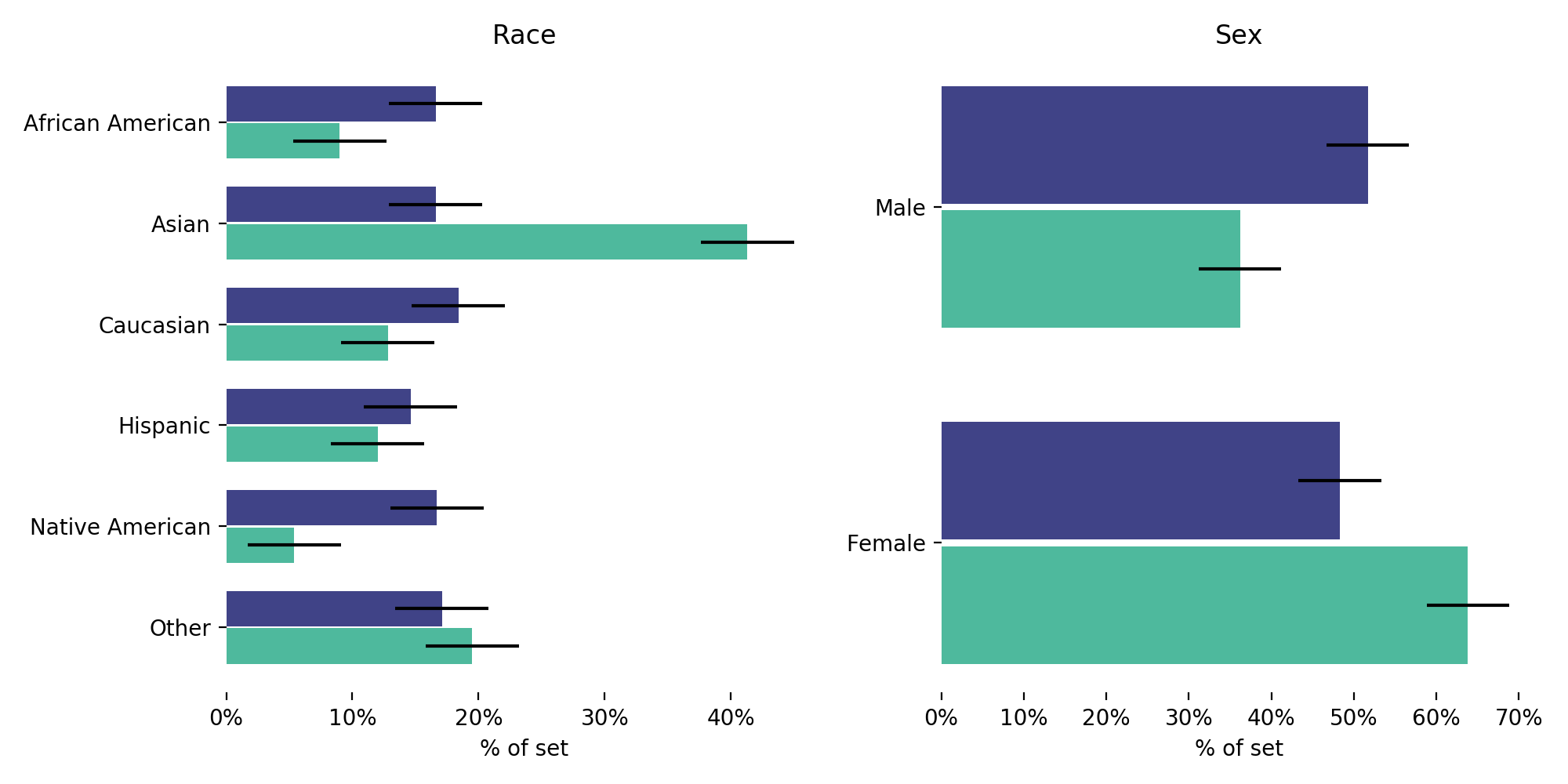}
    \caption{To locate unfairness in the binary classifier trained on the COMPAS data set, we apply LUCID to assess if the protected features have a uniform distribution in the canonical set. This canonical set was created with a learning rate of $0.1$ looping over $200$ epochs. To assess if it is balanced w.r.t. the protected features ``sex,'' and ``race,'' their distribution before (dark blue) and after (light green) inverse design is analyzed. The error bars indicate the variance of a uniform distribution with the respective number of categories. We learn that both the ``race'' and ``Sex,'' feature do not keep their uniform distribution after inverse design. This indicates a preference of the model for certain values of these features.}
    \label{fig:interpretcompas}
\end{figure}

To locate the unfairness in the classifier trained on the UCI Adult data set, we assess the distributions of the protected features “sex,” “race,” “marital status,” and  “relationship” in the canonical set (see Fig.~\ref{fig:interpret1}). We see that the ``race'' feature keeps its initial uniform distribution, and ``Sex,'' ``marital status,'' and ``relationship'' do not after the canonical inverse design. This indicates a preference of the model for certain values of these features. Additionally, three numerical features are analyzed: ``age,'' ``education level,'' and ``hours per week.'' All three distributions shift to larger values to achieve a positive output. In the canonical set for the classifier trained on the COMPAS data set both the ``race,'' and ``sex'' feature do not keep their initial uniform distribution after the canonical inverse design (see Fig.~\ref{fig:interpretcompas}). This indicates that women and Asian people are treated preferentially.

We compare the results of LUCID with two well--known traditional output-based notions of assessing fairness based on group membership: Statistical Parity and Equal Opportunity ~\cite{makhlouf2021applicability}. Statistical Parity holds when all subpopulations have an equal Positivity Rate (PR). This means that the same proportion of each subpopulation receives a positive output. Equal Opportunity holds when all subpopulations have an equal True Positive Rate (TPR). This implies that for each subpopulation the same rate of people who should receive a favorable output also receive this output.

\begin{table*}[t]
\caption{Positivity Rates and True Positive Rates of Subpopulations for the Protected Features in the UCI Adult Data Set.}
\label{tab:Fmetricsadult}
\begin{center}
\begin{small}
\begin{sc}
\begin{tabular}{c|ccccccc}
     & Male & \multicolumn{1}{c|}{Female}&White&  \shortstack{Asian Pac.\\Islander} & \shortstack{Amer. Indian\\Eskimo}&Black&Other\\
    PR & 24.7&\multicolumn{1}{c |}{8.0}&47.7&0.6&38.6&0.6&4.2\\
    TPR &60.0&\multicolumn{1}{c|}{54.6}&75.0&29.5&62.4&13.3&34.0\\
    &&&&&&&\\
      &Married & Divorced & \shortstack{Never\\married} & Separated & Widowed & \shortstack{Spouse\\absent} & \shortstack{Military\\spouse}\\
    PR & 38.8 & 4.3 & 1.6 & 3.4 & 5.0 & 5.7 & 42.9\\
    TPR & 63.4 & 36.5 & 28.5 & 33.3 & 46.5 & 41.7 & 100.0\\
    &&&&&&&\\
     &Wife & Own child&Husband &Not in family& Other relatives & Unmarried\\
    PR & 47.7&0.6&38.6&4.2&0.6&2.4\\
    TPR & 75.0&29.5&62.4&34.0&13.3&34.1\\
\end{tabular}
\end{sc}
\end{small}
\end{center}
\end{table*}

\begin{table*}[t]
\caption{Positivity Rates and True Positive Rates of Subpopulations for the Protected Features in the COMPAS Data Set.}
\label{tab:Fmetricscompas}
\begin{center}
\begin{sc}
\begin{tabular}{c|cccccccc}
     & Male & \multicolumn{1}{c|}{Female}&\shortstack{African\\ American} & Asian & Caucasian & Hispanic &\shortstack{Native \\American}&Other\\
     &&\multicolumn{1}{c|}{}&&&&&&\\
    PR &76.8&\multicolumn{1}{c |}{90.8}&45.3&100.0&75.6&75.0&20.0&93.7\\
    TPR &50.0&\multicolumn{1}{c|}{59.7}&30.2&71.4&51.6&50.0&20.0&52.4\\
\end{tabular}
\end{sc}
\end{center}
\end{table*}

In Table~\ref{tab:Fmetricsadult}, we show the PR and the TPR of the respective subpopulations of the four protected features for the UCI Adult data set. We see that none of the protected features adheres to either Demographic Parity or Equal Opportunity within an acceptable error margin. Interestingly, these biases differ from those found with LUCID. For example, for the ``race'' feature, the two fairness metrics indicate that the categories ``White'' and ``American Indian Eskimo'' have better rates and, therefore, a benefit. However, with LUCID, we do not find a preferred ``race'' category. The statistical metrics also find a bias in the ``martial status'' and ``relationship'' features. For ``relationship,'' the highest value of the two rates corresponds to the most preferred category in the canonical set. However, the second-highest value differs. The traditional metrics indicate ``Husband'' while the canonical set indicates ``Not in family.'' For the ``Marital status'' feature the highest and the second-highest values are exchanged, i.e., ``Married'' and ``Military spouse.'' In Table~\ref{tab:Fmetricscompas}, we present the PR and the TPR of the respective subpopulations of the two protected features for the COMPAS data set. Overall, the results of these fairness metrics are similar to the results of LUCID.


While LUCID confirms the results of the fairness metrics on the COMPAS data set, there is not always a one-to-one match on the UCI Adult data set. These differences may be the result of several distinct properties of both data sets. First, the UCI Adult data set contains a lot more variables which can encode the information embedded in the protected features. These ``proxy'' variables may be used by the model to make certain decisions that result in discriminatory outputs. Second, the output-based fairness metrics focus on the outcome, i.e., whether or not the output probability is above or below a certain threshold. In contrast, LUCID is an input-based analysis that aligns with the increasing focus on equality of treatment. The canonical inputs maximize the model's preferred output. LUCID thus looks for the inputs which do not simply lie above the threshold, but which are strongly preferred by the model. Finally, a lot of categories in the protected features do not contain a lot of data points (e.g., ``Military spouse'' in the UCI Adult data set). The PR and TPR of these categories are not statistically relevant, and therefore the results may differ from LUCID. Even, in the case that the results are similar, LUCID provides a sanity check and may detect potential overfitting on these few data points. Note that the statistical metrics can only be evaluated using a ground truth and do not consider the internal dynamics of the model.


\section{Conclusion}
To ensure that algorithms are used for the benefit of society and not against it, they should be accessible for transparent evaluation. Above all because decision-making algorithms can create, propagate, support, and automate bias in decision-making processes. Therefore, assessing whether the internal logic of algorithms is ``fair'' is essential to mitigate these biased algorithms, and promoting equality of treatment.

We introduce LUCID which generates a trained model's canonical set through gradient-based inverse design. We show that this set provides meaningful information about the treatment of protected features by analyzing if it is balanced with respect to said features. By unleashing the technique on two models, trained on the UCI Adult and COMPAS data sets respectively, we find that analyzing the canonical set exposes several biases of which some interestingly differ from those found in traditional output analyses.

LUCID is a useful addition to the toolbox of algorithmic fairness evaluation, as it can be implemented without access to the training data set or the output, and explains the internal workings of the algorithm while locating the unfairness. Further research includes applying it on other types of models such as decision trees, determining potential clustering of inputs in the canonical set, and performing a dynamical analysis to examine which features adjust faster than others.

\bibliography{Bibliography.bib}

\end{document}